\documentclass{bmvc2k}

\newcommand\blfootnote[1]{%
	\begingroup
	\renewcommand\thefootnote{}\footnote{#1}%
	\addtocounter{footnote}{-1}%
	\endgroup
}

\title{Enhanced 3D convolutional networks for crowd counting}

\addauthor{Zhikang Zou\textbf{\textsuperscript{$\ast$}}}{zhikangzou001@gmail.com}{1}
\addauthor{Huiliang Shao\textbf{\textsuperscript{$\ast$}}}{u201615039@hust.edu.cn}{1}
\addauthor{Xiaoye Qu}{xiaoye@hust.edu.cn}{1}
\addauthor{Wei Wei}{weiw@hust.edu.cn}{2}
\addauthor{Pan Zhou\textsuperscript{$\dagger$}}{panzhou@hust.edu.cn}{1}

\addinstitution{
 School of Electronic Information and Communications,\\
 Huazhong University of Science and Technology,\\
 Wuhan, China
}
\addinstitution{
 School of Computer Science and Technology,\\
 Huazhong University of Science and Technology,\\
 Wuhan, China
}

\runninghead{ZOU,SHAO}{Enhanced 3D convolutional networks for crowd counting}

\def\etal{\emph{et al}\bmvaOneDot}

\begin{document}

\maketitle

\blfootnote{\hspace{-0.5cm}$\ast$ Equal contribution \\ $\dagger$ Corresponding author}

\begin{abstract}
Recently, convolutional neural networks (CNNs) are the leading defacto method for crowd counting. However, when dealing with video datasets, CNN-based methods still process each video frame independently, thus ignoring the powerful temporal information between consecutive frames. In this work, we propose a novel architecture termed as "temporal channel-aware" (TCA) block, which achieves the capability of exploiting the temporal interdependencies among video sequences. Specifically, we incorporate 3D convolution kernels to encode local spatio-temporal features. Furthermore, the global contextual information is encoded into modulation weights which adaptively recalibrate channel-aware feature responses. With the local and global context combined, the proposed block enhances the discriminative ability of the feature representations and contributes to more precise results in diverse scenes. By stacking TCA blocks together, we obtain the deep trainable architecture called enhanced 3D convolutional networks (E3D). The experiments on three benchmark datasets show that the proposed method delivers state-of-the-art performance. To verify the generality, an extended experiment is conducted on a vehicle dataset TRANCOS and our approach beats previous methods by large margins.
\end{abstract}

%-------------------------------------------------------------------------
\section{Introduction}
\label{sec:intro}
Crowd counting algorithms aim to produce an accurate estimation of the true number of individuals in a still image or a video. It has drawn much attention due to the important geo-political and civic applications in video surveillance, traffic control, and abnormally detection.
Moreover, some crowd counting methods with great generality can be extended to other applications, such as automobile counting at traffic jams, cell or bacteria counting from microscope images and animal estimation in wild scenes. However, it's still a challenging vision task to obtain accurate individual number because of severe occlusion, diverse distribution and perspective distortion.
% and cell or bacteria counts from microscope images.

Recently, researchers have leveraged Convolutional Neural Networks (CNNs) for an accurate crowd density map generation \cite{7298684,8497050,Sam_2018_CVPR,Liu_2018_CVPR_de}. Some works focus on explicitly incorporating multi-scale information based on multi-column architectures \cite{Zhang_2016_CVPR,Sam_2017_CVPR}. They use different filter sizes for different columns which are adaptive to the scale variation in crowd size. Instead, FCN-7c \cite{kang2018crowd} feeds an image pyramid of the input image into a single column network in order to alleviate heavy computational overhead. Though these methods have made significant progress, they are restricted by capturing the informative representations with local receptive fields, which isolates the pixels from the global scene context. In CP-CNN \cite{Sindagi_2017_ICCV}, Contextual Pyramid CNNs are proposed to explicitly incorporate global and local contextual information of crowd images which are fused with high-dimensional feature maps to generate accurate density maps. However, they need to train two additional networks to evaluate
the context of crowds, which suffers from high computation complexity. Moreover, existing CNN-based methods are faced with inherent algorithmic weaknesses: when dealing with video sequences, they still regard the data as single still images, thus ignoring the temporal information stored in neighbouring frames. To exploit the strong correlation in video data, Xiong \etal \cite{Xiong_2017_ICCV} propose a variant of LSTM in order to process the sequence of images into density maps. However, that LSTM is difficult to train hinders the wide application of the proposed method.

To practically resolve these problems, we propose a novel temporal channel-aware (TCA) block to utilize the correspondence among video sequences and capture global feature statistics simultaneously. Motivated by the achievement of 3D CNN in action recognition \cite{Tran_2015_ICCV}, we employ 3D convolutions in the proposed block to encode spatio-temporal features for videos, especially improving the representation ability in temporal dimension. Besides, global contextual information is transformed into modulation weights which adaptively highlight the useful features by rescaling each channel-aware feature response. We also leverage short skip connections to ease the training of the model. Therefore, we could stack several TCA blocks together to form very deep network, named as enhanced 3D convolutional networks (E3D). Extensive experiments on three benchmark datasets (inculding WorldExpo'10 \cite{7298684}, UCSD \cite{4587569}  and MALL \cite{Chen_2012_BMVC}) show that the proposed E3D yields significant improvement over recent state-of-the-art methods. Furthermore, we evaluate E3D on a vehicle dataset TRANCOS \cite{10.1007/978-3-319-19390-8_48} to demonstrate the generality for counting other objects.

To summarize, we make the following contributions:
% \newline
\begin{itemize}
\item To the best of our knowledge, it's the first attempt to adopt 3D convolution for crowd counting. By introducing 3D kernels, the model is capable of capturing the temporal and spatial information simultaneously, thereby boosting the performance on video datasets.
% \newline
\item We design a novel temporal channel-aware (TCA) block to incorporate both local and global spatio-temporal information. By applying the proposed block into the enhanced 3D convolutional networks (E3D), the network achieves tremendous improvement in strengthening the discriminative ability of feature representations.
% \newline
\item We conduct a throughout study on the number of frames sent to network, the number of TCA blocks and the components of the whole architecture. 
\end{itemize}

\section{Related Work}
Plenty of algorithms have been proposed for crowd counting to solve related real world problems \cite{Liu_2018_CVPR,Ranjan_2018_ECCV,Shi_2018_CVPR,onoro2018learning}. 
% The previous literature can be divided into two categories: detection-based approaches and regression-based approaches.
Most of the early researchers focus on detection-based framework using a moving-window-like detector to estimate the number of individuals. These methods require well-trained classifiers to extract low-level features from a full body such as Haar Wavelets \cite{Viola2004} and HOG \cite{dalal:inria-00548512}. However, for crowded scenarios, objects are highly occluded and difficult to detect. To tackle this problem, researchers have attempted to detect particular body parts instead of the whole body to estimate the count \cite{Wu_2007_ICCV,4761705}. For instance, Li \etal \cite{4761705} incorporate a foreground segmentation algorithm and a HOG-based head-shoulder detection algorithm to detect heads, thus implementing crowd scenes analysis.

Although the part-based detection methods alleviate the problem of occlusion, they perform poorly on extremely congested scenes and high background clutter scenes. Researchers make an effort to deploy regression-based approaches which learn a mapping between extracted features from cropped images patches and their count or density \cite{4587569,Chen_2012_BMVC}. Moreover, some methods \cite{5459191,5384566} leverage spatial or depth information and take approach of segmentation methods to filter the background region, thereby regressing count numbers only on foreground segments. These methods are sensitive to different crowd density and have addressed the problem of severe occlusion in dense crowds. 

% In recent methods, instead of directly regressing the total crowd count, Lempitsky et al.\cite{NIPS2010_4043} propose to learn a linear mapping between local patch features and corresponding object density maps. Pham et al.\cite{7410729}  proposes a method to learn a non-linear mapping via random forest regression after observing the difficulty of learning a linear mapping.

Recently, CNN-based approaches have become ubiquitous owing to its success in a large number of computer vision tasks \cite{Miao_2018_CVPR,DBLP:journals/corr/abs-1711-11248,Zhang_2018_CVPR}. Many researchers have shifted their attention towards CNN-based methods, which have achieved significant improvements over previous methods of crowd counting. Zhang \etal \cite{Zhang_2016_CVPR} propose a multi-column based architecture to tackle the large scale variations in crowd scenes. Similarly, Onoro \etal \cite{10.1007/978-3-319-46478-7_38} develope a scale-aware counting model called Hydra CNN for object estimation. Sam \etal \cite{Sam_2017_CVPR} use a switch layer to select the most optimal regressor for the particular input patches. In order to pursue the quality of the density maps, Sindagi \etal \cite{Sindagi_2017_ICCV} propose contextual pyramid networks to generate high-quality density maps by explicitly combining global and local contextual information at the expense of complicated structures. However, multi-column structures are much more difficult to train because each subnet may have different loss surfaces. Therefore, SCNet \cite{Wang_2018_BMVC} makes a balance between pixel-wise estimation and computational costs by designing a single-column network. Further, Li \etal \cite{Li_2018_CVPR} incorporate dilated convolutions to aggregate multi-scale contextual information.

% Regrettably, most of the previous models only pursue the accuracy of the crowd counts regardless of the quality of density map. \cite{Sindagi_2017_ICCV} propose contextual pyramid network for generating high-quality crowd density by explicitly combining global and local contextual information at the expense of complicated network. \cite{Wang_2018_BMVC} make a balance between pixel-wise estimation and computational costs by designing a single-column network termed as SCNet. Li \etal \cite{Li_2018_CVPR} propose a dilated convolution to aggregate and multi-scale contextual information. \cite{kang2018crowd} adaptively fuse the predictions from different scales by softly selecting a suitable scale for each pixels. 

Despite the promising results, all the methods mentioned above neglect the otherwise powerful temporal information when dealing with video data. To deal with this problem, some methods \cite{Xiong_2017_ICCV,DBLP:journals/corr/ZhangWCM17aa} attempt to utilize variants of LSTM to access long-range temporal dependencies. However, the complex LSTM architecture indicates the requirement of heavy computational costs and difficulty of training relevant parameters. Instead, we introduce 3D convolutions to exploit temporal information among videos in this paper. A novel architecture termed as "temporal channel-aware" (TCA) block is designed to capture temporal interdependencies and encode global contextual information simultaneously.

%-------------------------------------------------------------------------
\section{Our approach}
Existing CNN-based crowd models mostly fail to take the temporal information into account for those images captured from video datasets. To overcome this issue, our solution is to adopt enhanced 3D convolutional network stacked by temporal channel-aware block to capture the strong temporal correlation.

\begin{figure}[ht]
\centering
\includegraphics[width=12cm]{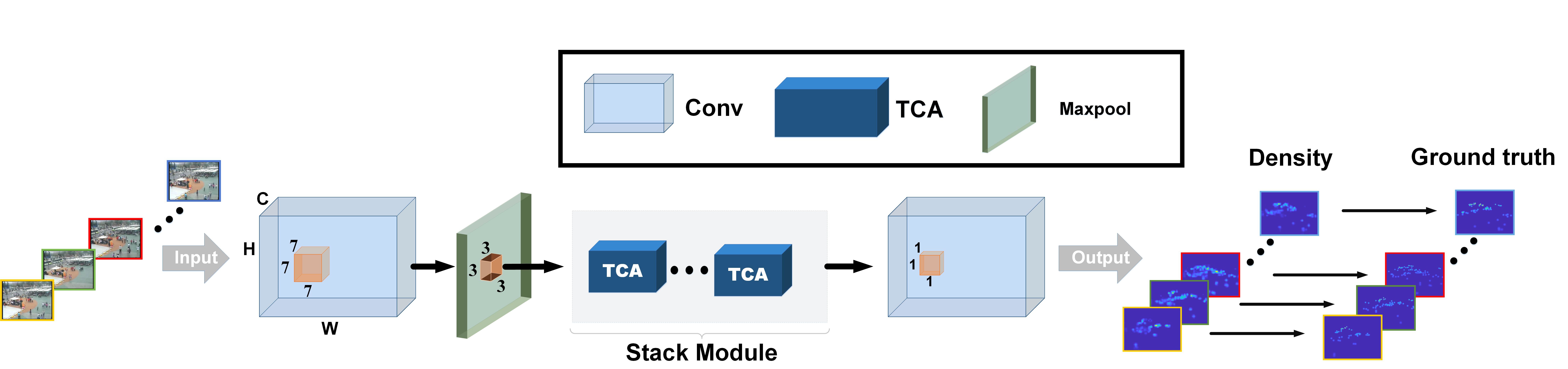}
\caption{Network architecture of our enhanced 3D convolutional network.}
\label{whole}
\vspace{-0.2cm}
\end{figure}

\subsection{Backbone architecture}
The architecture of the proposed E3D network is shown in Fig. \ref{whole}. Given multiple successive frames stacked as inputs, this sequence is first processed by a 3D convolution with a kernel size of 7x7x7 and stride 1x2x2. With the help of the sliding convolution operation among the time dimension, the temporal correlation between neighbouring input frames is captured. Then a max-pooling with pooling size of 3x3x3 and stride 1x1x1 is applied to downsample the extracted feature maps. To further explore the spatio-temporal information, we stack eight TCA blocks which can analyze the features from both local and global perspectives. The number of TCA blocks is chosen according to the performance on all testing datasets.
There are two different types of TCA blocks used in the stream with slight nuance as one type changes the stride of the first convolution to 1x2x2 in order to downsample features in the spatial dimension while the others stay 1x1x1. We stack these two TCA blocks alternately and make sure that the output of the stack module is 1/16 of the input size. The detailed description of the TCA block will be introduced in the next section. After this stage, we use a 3D convolution with a kernel size of 1x1x1 to match the channel of output feature maps with the ground-truth density maps. Meanwhile, we scale the ground-truth density maps using bilinear interpolation with the factor of 1/16 to match the size of the final density maps. The proposed E3D is a fully convolutional network to accept inputs of arbitrary sizes and can be optimized via an end-to-end training scheme.

\subsection{TCA Block}
\setlength{\abovecaptionskip}{-2pt}
\begin{figure}[ht]
\centering
\includegraphics[width=12cm]{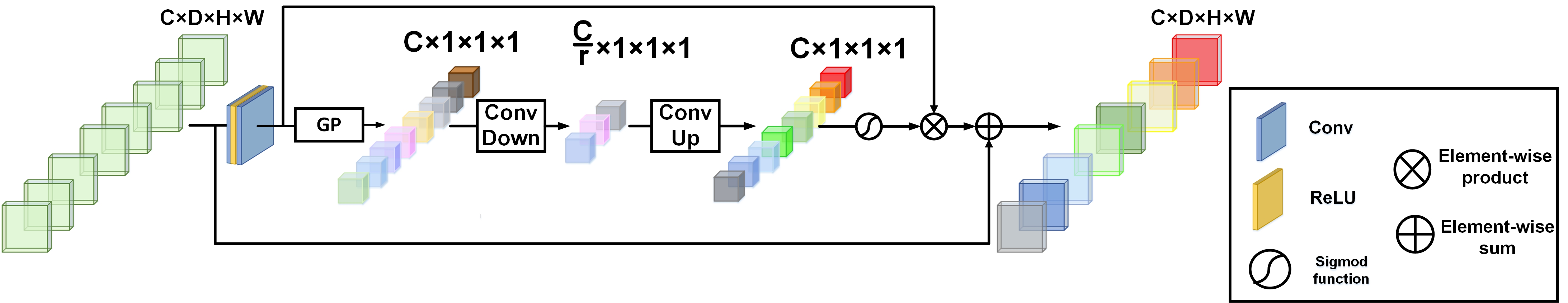}
\caption{Temporal channel-aware block}
\label{TCA}
\end{figure}
% \newline 
\noindent The critical component of our architecture is the TCA block, as is depicted in Fig. \ref{TCA}. This block can be divided into two branches, namely the mainstream and the shortcut branch. The mainstream branch deals with the feature maps and reconstructs the channel-wise feature response, while the shortcut branch here is to leverage the effectiveness of the residual learning for effective training whose spirit is similar to ResNet \cite{he2016deep}. Specifically, in the mainstream branch, the feature maps $X$ input to the block first pass two 3D convolutional layers. In this transformation, the number of channels remains unchanged. We use $c$ and $c'$ to represent the number of input and output channels, respectively. For simplicity, bias terms are omitted and each 3D convolutional transformation can be formulated as:
\begin{equation}
o_{c'}=k_{c'}\ast X = \sum_{i=1}^{c} k_{c'}^{i} \ast x^{i},
\end{equation}
where * denotes convolution, ${k_{c'}=[ k_{c'}^{1}, k_{c'}^{2}}$, ${\cdots k_{c'}^{c} ]}$ is the set of learned filter kernels and ${k_{c'}^{i}}$ is a 3D spatial kernel for a single input channel, ${X=[ x^{1},x^{2},\cdots x^{c}]}$ are input feature maps.
Because $c'$ is equal to $c$, thus we still use $c$ for following notations. After the process of two convolutions and a relu, the output features $O$ can be denoted as ${O = \left [ o_{1}, o_{2}, \cdots o_{c}\right ] }$. Global contextual information is expected to be fused into both spatial and temporal dimensions because of the specific informative features in each channel, which indicates that it's inapposite to treat all channels with equality.
Therefore, we send the output features to a channel descriptor by aggregating feature maps across their spatio-temporal dimensions. The channel descriptor is produced by global average pooling to generate channel-wise weights so that our network can selectively increase its sensitivity to useful informative features which can be effectively exploited by subsequent transformations. Formally, channel-wise means $v_{c}$ are generated by shrinking $O$ through spatiotemporal dimensions ${D \times H \times W}$: 
\begin{equation}
    v_{c}=\frac{1}{D\times H \times W}\sum_{l=1}^{D}\sum_{m=1}^{H}\sum_{n=1}^{W} o_{c}\left ( l,m,n \right ).
\end{equation}
To further exploit the channel dependencies, we use a dimensionality-reduction convolution layer followed by a dimensionality-increasing convolution layer to automatically learn the subtle interaction relationships between channels. We then utilize a sigmoid activation to normalize the weights of multiple channels. This procedure can be defined as:
\begin{equation}
    u=\sigma(W_{2}\delta (W_{1}v)),
    \quad {{v}=\left [ {v}_{1},{v}_{2},\cdots, {v}_{c} \right ]},
\end{equation}
where ${ {u}=\left [ {u}_{1},{u}_{2},\cdots, {u}_{c} \right ]}$ is the normalized weight, ${\delta}$ refers to the ReLU function, ${\sigma}$ refers to the sigmoid function, 
$W_{1} \in {R^{\tilde{C} \times C}}$, $W_{2} \in {R^{C \times \tilde{C}}}$ are the convolutions, $\tilde{C} = \frac{C}{r}$ and $r$ is the reduction ratio. In this paper, $r$ is set to 4. With the normalized channel weights, the channel information in the TCA block is adaptively rescaled. The output of the mainstream branch can be achieved by channel-wise multiplication between the feature map $o_{c} \in R^{D \times H \times W}$ and the normalized channel weight $u_{c}$. 
The process of producing output can be formulated as:
\begin{equation}
    {\tilde{O}=\left [ \tilde{o}_{1},\tilde{o}_{2},\cdots, \tilde{o}_{c} \right ]},\quad
    \tilde{o_{c}}=o_{c}\cdot u_{c}.
\end{equation}
Finally, by integrating information from both branches, we can get the final output of TCA block as:
\begin{equation}
\tilde{X}=X+\tilde{O}.
\end{equation}

\subsection{TCA-2D: a degenerate variant of TCA block}
To get a further understanding of the effectiveness of exploiting temporal information, we propose a degenerate variant of TCA block in a 2D version called TCA-2D, which is stacked to form enhanced 2D convolutional networks termed as E2D. Except for the downsampling layer, we replace all the 3D kernels with the corresponding 2D ones. With regard to the downsampling layers with stride 1x2x2, we use 2x2 to replace the original stride. 

In the experiments to be reported in the next section, whenever the dataset consists of images not captured from the same video sequences, the E3D will not come into effect but only E2D will be employed.

\subsection{Ground Truth Generation}
Following the method of generating density maps in \cite{7298684}, we generate the ground truth by blurring each head annotations via a Gaussian kernel which is normalized to sum to one. Therefore, the total sum of the density map equals to the actual crowd counts. The ground truth $G$ is given as follow:
\begin{equation}\label{eq1}
 G(x)=\sum_{i=1}^{N}\delta(x-x{_{i}})\ast G{_{\sigma }}(x),
\end{equation}
where $N$ is the total number of the individuals and $G_{\sigma}(x)$ represents 2D Gaussian kernels. Considering the negative effect of perspective distortion to some extent, we employ the geometry-adaptive kernels \cite{Zhang_2016_CVPR} to process the datasets lack of geometry information. The geometry-adaptive kernels are defined as:
\begin{equation}\label{eq2}
F(x)=\sum_{i=1}^{N}\delta (x-x_{i})\times G_{\sigma_{i} }(x), \qquad \sigma_{i}=\beta \bar{d^i}.
\end{equation}
For each head x$_{i}$, we use ${\bar{d^i}}$ to indicate the average distance of $k$ nearest neighbours. ${\delta (x-x_{i})}$ is convolved with a Gaussian kernel with standard deviation parameter ${\sigma_{i}}$  where $x$ is the position of pixel in each image. In the experiment, the ratio ${\beta}$ is set to 0.3 and $k$ is 3.

\section{Experiment}
\begin{figure}[ht]
\centering
\includegraphics[width=12cm]{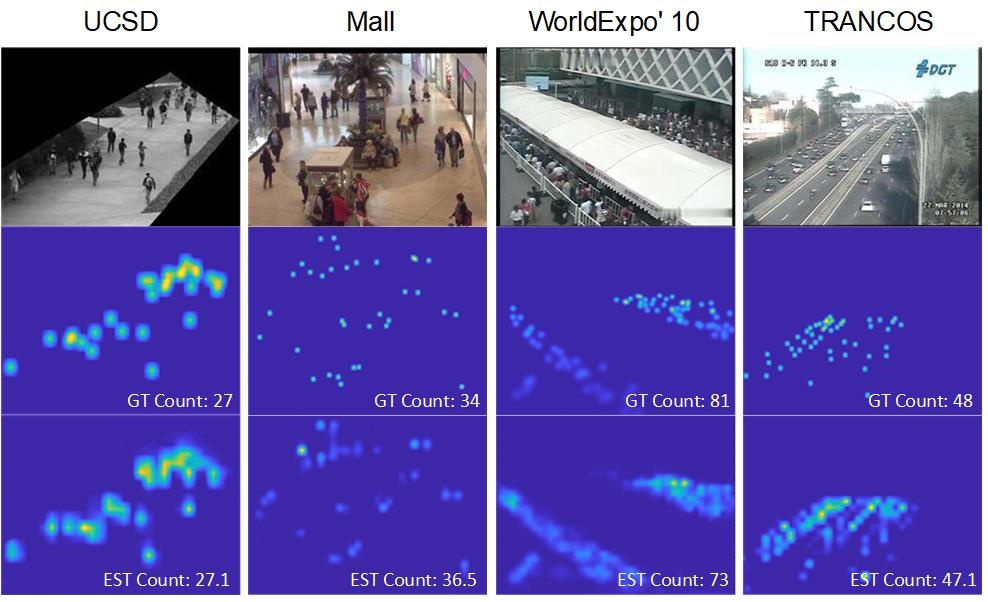}
\caption{Visualization of estimated density maps on four benchmark datasets by the proposed enhanced 3D convolutional networks.}
\label{visualization}
%\vspace{-0.05cm}
\end{figure}

We demonstrate the effectiveness of the proposed E3D model on three popular video datasets as well as the vehicle dataset TRANCOS. Some statistics of these datasets and the corresponding kernels we use are summarized in Table \ref{statistics}. Besides, ablation studies are conducted on the UCSD dataset to analyze the impact of the number of video frames sent to the network, the effect of the number of the stacked TCA blocks and the capability of each component in our network. Qualitative results are visualized in Fig. \ref{visualization}.

\vspace{-0.1cm}
\begin{table}[htbp]
%\begin{center}
\setlength{\abovecaptionskip}{2pt}
\resizebox{\textwidth}{10mm}{
\begin{tabular}{|l|l|l|l|c|c|c|c|c|c|}
\hline
Dataset & Resolution & Color & Num & FPS & Max & Min & Average & Total & Generating method \\
\hline\hline
UCSD \cite{4587569}& 238 $\times$ 158 & Grey & 2000 & 10 & 46 & 11 & 24.9 &49885 & Fixed kernel: $\sigma$=4\\
\hline
Mall \cite{Chen_2012_BMVC}& 640 $\times$ 480 & RGB & 2000 & < 2 & 53 & 11 & 31.2 & 62315 & Geometry-adaptive kernels \\
\hline
WorldExpo'10 \cite{7298684}& 720 $\times$ 572& RGB & 3980 & 50 & 253 & 1 & 50.2 & 199923 & Fixed kernel: $\sigma$=3\\
\hline
TRANCOS \cite{10.1007/978-3-319-19390-8_48}& 640 $\times$ 480 & RGB &  1244 &- & -&-&37.6 & 46796 & Fixed kernel: $\sigma$=4 \\
\hline
\end{tabular}}
\caption{Statistics of the four datasets}
\label{statistics}
%\end{center}
\end{table}
\vspace{-0.7cm}

\subsection{Evaluation metrics}
The widely used {\em mean absolute error} (MAE) and the {\em mean squared error} (MSE) are adopted to evaluate the performance of different methods. The MAE and MSE are defined as follows:
\begin{equation}\label{eq4}
\text{MAE}=\frac{1}{N} \sum_{i=1}^{N}\left | q_{i}-\hat{q_{i}} \right |, \qquad \text{MSE}=\sqrt{\frac{1}{N}\sum_{i=1}^{N}(q_{i}-\hat{q_{i}})^2}.
\end{equation}
Here, $N$ represents the total number of frames in the testing datasets, ${q_{i}}$ and ${\hat{q_{i}}}$ are the ground truth and the estimated count, respectively. ${\hat{q_{i}}}$ is calculated by summing up the estimated density map over the entire image.

\subsection{Results}
\textbf{UCSD.} The UCSD crowd counting dataset \cite{4587569} consists of 2000 video frames of pedestrians on a walkway of the UCSD campus captured by a stationary camera. The video was recorded at 10fps with dimension 238 ${\times}$ 158. A region of interest(ROI) is provided for the scene in the dataset so that all frames and corresponding ground truth are masked with ROI. In the final output feature map, the intensities of pixels out of ROI is also set to zero, thereby constraining the error to the ROI areas to backpropagate during training. Following the setting in \cite{4587569}, we use frame 601-1400 as the training data and the remaining 1200 frames as test data. We adopt a fixed spread Gaussian to generate ground truth density maps for training the network as the crowd is relatively sparse. Considering the fact that the resolution of each frame is small and fixed, each image is resized to two times the original size before it is sent to the network. We send 16 frames to the network at a time. The results of the different methods are shown in Table \ref{ucsd}. Compared with other state-of-the-art approaches, our method achieves the best result, which can be regarded as a verification that temporal information can boost the performance for this dataset.
\begin{table}[!htp]
\setlength{\abovecaptionskip}{2pt}
\begin{minipage}[a]{0.45\textwidth}
\centering
\resizebox{\textwidth}{16mm}{
\begin{tabular}{|l|c|c|}
\hline
Method & MAE & MSE \\
\hline\hline
% Gaussian process regression & 2.24 & 7.97 \\
% \hline
Cross-Scene  \cite{7298684} & 1.60 & 3.31 \\
\hline
MCNN \cite{Zhang_2016_CVPR} & 1.07 & 1.35\\
\hline
Bidirectional ConvLSTM \cite{Xiong_2017_ICCV} & 1.13 & 1.43\\
\hline
Switching-CNN \cite{Sam_2017_CVPR}& 1.62 & 2.10\\
\hline
% FCN-5c-3s \cite{kang2018crowd} & 1.16 & 2.29 \\
% \hline
CSRNet \cite{Li_2018_CVPR}& 1.16 & 1.47\\
\hline
SANet \cite{Cao_2018_ECCV} & 1.02 & 1.29 \\
\hline
E3D (ours)& \textbf{0.93} & \textbf{1.17}\\
\hline
\end{tabular}}
\caption{Performance on UCSD dataset}
\label{ucsd}
\end{minipage}
\qquad
\begin{minipage}[a]{0.45\textwidth}
\centering
\resizebox{\textwidth}{16mm}{
\begin{tabular}{|l|c|c|}
\hline
Method & MAE & MSE \\
\hline \hline
% Gaussian process regression & 3.72 & 20.1 \\
% \hline
Ridge regression \cite{Chen_2012_BMVC} & 3.59 & 19.1 \\
\hline
Count forest \cite{7410729}& 2.50 & 10.0\\
% \hline
% MoCNN \cite{}& 2.75 & 13.40\\
% \hline
\hline
Weighted VLAD \cite{7778134} & 2.41 & 9.12 \\
\hline
Exemplary Density \cite{7533041} & 1.82 & 2.74 \\
\hline
MCNN \cite{Zhang_2016_CVPR} & 2.24 & 8.5 \\
\hline
Bidirectional ConvLSTM \cite{Xiong_2017_ICCV} & 2.10 & 7.6\\
\hline
E3D (ours)& \textbf{1.64} & \textbf{2.13}\\
\hline
\end{tabular}}
\caption{Performance on Mall dataset}
\label{mall}
%\end{subtable}
\end{minipage}
\end{table}
\newline \textbf{Mall.} The mall dataset \cite{Chen_2012_BMVC} was provided by Chen {\em et al.} for crowd counting. It was collected from a public accessible webcam in a shopping mall. The video contains 2000 annotated frames of over 60000 pedestrians with their head positions labeled. The ROI is also provided in the dataset. We use the first 800 frames for training and the remaining 1200 frames for testing. With more challenging lighting conditions and glass surface reflection, it's difficult to find the underlying relationship between the head size and density map. Thus geometry-adaptive kernels are applied to generate the density map. Also, 16 frames are sent to the network simultaneously. We perform a comparison against previous methods and our method achieves state-of-the-art performance with respect to both MAE and MSE. The results are shown in Table \ref{mall}, which also verifies the effectiveness of the powerful temporal information between recurrent frames.
\newline \newline \textbf{WorldExpo'10.} The WorldExpo'10 dataset \cite{7298684} is made up of 3980 annotated frames from 1132 video sequences captured by 108 different surveillance cameras. This dataset is split into a training set of 3380 frames collected by 103 cameras and a testing set of 600 frames from 5 different scenes. The region of interest (ROI) are provided for these five test scenes. Each frame and its dot maps are masked with ROI during processing. We still use 16 frames as inputs and the MAE metric for evaluation. As shown in Table \ref{world}, the proposed E3D achieves the best accuracy in 4 out of 5 scenes and delivers the lowest average MAE compared with previous methods.
% Also, the datasets provide perspective maps which specify the number of the image covering one square meter at realistic location. To be consistent with the work of[][], we generate the labeled density map using perspective maps that adaptively choose the spread of Gaussian.
\newline 
\newline
\noindent\textbf{TRANCOS.}  In addition to counting pedestrians, an extended experiment is conducted on the vehicle dataset TRANCOS \cite{10.1007/978-3-319-19390-8_48} to present the generality of our model. It consists of 1244 images of different congested traffic scenes with a total of 46796 vehicles annotated. Different from crowd counting datasets, TRANCOS contains multiple scenes from different video sequences, which indicates intercepted frames are not consecutive. Since there is no temporal correlation between them, we can not make use of the advantage of 3D convolutions. Therefore, we degenerate the 3D model into a 2D version named E2D, which is made up of TCA-2D blocks. Strictly following the setting in \cite{10.1007/978-3-319-19390-8_48}, we adopt the Grid Average Mean absolute Error (GAME) metric, which is:
\begin{equation}\label{eq7}
\text{GAME}(L)=\frac{1}{N}\sum_{n=1}^{N}(\sum_{l=1}^{4^{L}}\left | C_{I_{n}}^{l}-C_{I_{n}}^{l_{GT}}\right |),
\end{equation}
where N is the number of test images, ${C_{I_{n}}^{l}}$ is the estimated count of image n within region $l$ and ${C_{I_{n}}^{l_{GT}}}$ is the corresponding ground truth result. The GAME metric aims at subdividing the image into ${4^L}$ non-overlapping region and evaluating the accuracy of the estimated position. When L=0, the GAME is equivalent to MAE. We compare our approach with five previous methods in Table \ref{TRANCOS} and achieve a
significant improvement in four different GAME metrics. This illustrates that our model can still obtain robust results in the absence of temporal information.
% Obviously, we giving the state-of-the-art results across all levels, which demonstrates the robustness of the degenerate variant of E3D and accuracy of capturing the crowd distribution.

\begin{table}[!htp]
\setlength{\abovecaptionskip}{2pt}
\renewcommand\arraystretch{1.2}
\begin{minipage}[m]{0.47\textwidth}
\centering
\resizebox{\linewidth}{!}{
\begin{tabular}{|l|c|c|c|c|c|c|}
\hline
Method & S1 & S2 & S3 & S4 & S5 & Avg \\
\hline\hline
Cross-Scene \cite{7298684} & 9.8 &14.1 & 14.3 & 22.2 & 3.7 & 12.9\\
\hline
MCNN \cite{Zhang_2016_CVPR} & 3.4 & 20.6 & 12.9 & 13.0 & 8.1 & 11.6\\
% \hline
% Bidirectional ConvLSTM \cite{Xiong_2017_ICCV}& 6.8 & 14.5 & 14.9 & 13.5 & \textbf{3.1} & 10.6\\
\hline
Switching-CNN \cite{Sam_2017_CVPR}& 4.4 & 15.7 & 10.0 & 11.0 & 5.9 & 9.6\\
\hline
CP-CNN \cite{Sindagi_2017_ICCV} & 2.9 & 14.7 & 10.5 & 10.4 & 5.8 & 8.86\\
\hline
% CSRNet \cite{Li_2018_CVPR} & 2.9 & \textbf{11.5} & \textbf{8.6} & 16.6 & 3.4 & 8.6 \\
% \hline
SCNet \cite{Wang_2018_BMVC} & \textbf{1.8} & 9.6 & 14.2 & 13.3 & 3.2 & 8.4 \\
\hline
E3D (ours)& 2.8 &12.5 & 12.9& \textbf{10.2} & \textbf{3.2} & \textbf{8.32} \\
\hline
\end{tabular}}
\caption{Performance on WorldEXpo'10}
\label{world}
\end{minipage}
\hspace{0.15cm} 
\begin{minipage}[m]{0.49\textwidth}
\centering
\resizebox{\linewidth}{!}{
\begin{tabular}{|l|c|c|c|c|}
\hline
Method & GAME0 & GAME1 & GAME2 & GAME3 \\
\hline\hline
% Cross-Scene \cite{7298684} & 11.24 &12.36 &14.51 &18.67 \\
% \hline
% Lempitsky {\em et al.} \cite{NIPS2010_4043}& 13.76 & 16.72 & 20.72 & 24.36 \\
% \hline
Hydra 3s \cite{10.1007/978-3-319-46478-7_38}& 10.99 & 13.75 & 16.69 & 19.32 \\
\hline
FCN-HA \cite{DBLP:journals/corr/ZhangWCM17aa} & 4.21 & - & - & - \\
\hline
AMDCN \cite{Deb_2018_CVPR_Workshops} & 9.77 & 13.16 & 15.00 & 15.87\\
\hline
FCNN-skip\cite{8360001} & 4.61 & 8.39 & 11.08 & 16.10\\
\hline
CSRNet \cite{Li_2018_CVPR}& 3.56 & 5.49 & 8.57 & 15.04\\
\hline
E2D (ours) & \textbf{2.88} & \textbf{4.81} & \textbf{7.77} & \textbf{12.47}\\
\hline
\end{tabular}}
\makeatletter\def\@captype{table}\makeatother\caption{Performance on TRANCOS}
\label{TRANCOS}
\end{minipage}
\vspace{-0.3cm}
\end{table}

\begin{table}[!htp]
\renewcommand\arraystretch{1.2}
\setlength{\abovecaptionskip}{2pt}
\centering
\resizebox{\linewidth}{!}{
\begin{tabular}{|c|c|c||c|c|c||c|c|c|}
\hline
Method & MAE & MSE & Frame length & MAE & MSE & TCA numbers & MAE & MSE\\
% E2D (w/o gc) & 1.10 & 1.40 \\
\hline\hline
E2D (w/o gc) & 1.10 & 1.40 & 4 & 1.26 & 2.79 & 4 & 1.18 & 1.48\\
\hline
E2D & 1.00 & 1.31 & 8 & 1.27 & 2.32 & 6 & 1.12 & 1.45\\
\hline
E3D (w/o gc) & 1.00 & 1.33 & 12 & 1.08 & 1.40 & 8 & \textbf{0.93} & \textbf{1.13}\\
\hline
E3D & \textbf{0.93} & \textbf{1.13} & 16 & \textbf{0.93} & \textbf{1.13} & 10 & 1.02 & 1.35 \\
\hline
\end{tabular}}
\caption{The ablations on UCSD about the capability of each component in our network, the impact of the number of video frames sent to the network and the effect of the number of the stacked TCA blocks.}
\label{ablation}
\end{table}

% \begin{table}[!htp]
% \renewcommand\arraystretch{1.2}
% \setlength{\abovecaptionskip}{2pt}
% \begin{minipage}[m]{0.5\textwidth}
% \centering
% \resizebox{\linewidth}{!}{
% \begin{tabular}{|c|c|c|c|c|}
% \hline
% Method & E2D (w/o gc) & E2D & E3D (w/o gc) & E3D \\
% % E2D (w/o gc) & 1.10 & 1.40 \\
% \hline\hline
% MAE & 1.10 & 1.00 & 1.00 & \textbf{0.93}\\
% \hline
% MSE & 1.40 & 1.31 & 1.33 & \textbf{1.13}\\
% \hline
% \end{tabular}}
% \caption{Component analysis}
% \label{component}
% \end{minipage}
% \hspace{0.15cm}
% \begin{minipage}[m]{0.38\textwidth}
% \centering
% \resizebox{\linewidth}{!}{
% \begin{tabular}{|c|c|c|c|c|}
% \hline
% Frame length & 4 & 8 & 12 & 16\\
% \hline \hline
% MAE  & 1.26 & 1.27 & 1.08 & \textbf{0.93} \\
% \hline
% MSE & 2.79 & 2.32 & 1.4 & \textbf{1.13} \\
% \hline
% \end{tabular}}
% \caption{Compact of frame length}
% \label{clip}
% %\end{subtable}
% \end{minipage}
% \vspace{-0.3cm}
% \end{table}

\subsection{Ablation study}
To have more insights into our proposed method, we conduct ablation studies on UCSD datasets for its representative temporal information.

\textsl{\textbf{1) Component analysis:}} Our first study is to investigate the capability of each component in the proposed E3D and the results are listed in Table \ref{ablation} (left). The first method E2D (w/o gc) means that we remove the mainstream branch (the global context branch) in each TCA-2D block of the E2D, the same goes for E3D (w/o gc) based on E3D. From the table, we could see that incorporating global context can reduce the MAE
from 1.10 (E2D (w/o gc)) to 1.00 (E2D) or from 1.00 (E3D (w/o gc)) to 0.93 (E3D), which demonstrates its effectiveness in improving the performance of the proposed model. Besides, the performance discrepancy between E2D (w/o gc) and E3D (w/o gc) or E2D and E3D illustrates that it is useful to exploit the temporal information in video sequences, which supports our justification for the use of the 3D convolutions. 

\textsl{\textbf{2) Frame length:}} Given the benefits of temporal convolutions above, it is interesting to study the impact of the number of the frames sent to the network on the final performance. As shown in Table \ref{ablation} (middle), we gradually increase the number of frames at intervals of 4. It is obvious that the results are comparative in the case of frames 4 and 8 (MAE 1.26 vs 1.27). However, the network achieves significant performance improvement when the number of frames is 12 or 16. It is because the UCSD dataset is recorded at 10fps. There is little temporal information to make use of when frame length is less than 10. This may suggest that our model benefits from the increase of the number of frames sent to the network. Considering the limited computing resources, we finally set the frame length to 16.

\textsl{\textbf{3) TCA numbers:}} The proposed method is composed of several TCA blocks, and it is necessary to analyze the effect of the number of TCA blocks on the final performance. We gradually increase the number of TCA blocks at intervals 2 shown in Table \ref{ablation} (right). As is mentioned above, there are two types of TCA blocks stacked alternately to make up the whole architecture. The difference between them is whether there exists downsampling operation. When changing the number of TCA blocks, we only add or remove blocks without downsampling. Therefore, the output size of the network is maintained at 1/16 of the input resolution. It is obvious that the network delivers the best performance when the number of blocks equals to 8.

\section{Conclusion}
In this paper, we propose a novel block named temporal channel-aware (TCA) block, which not only captures the temporal dependencies in video sequences, but also combines global context information with local spatio-temporal features to boost the accuracy for crowd counting. By stacking the TCA blocks to form the enhanced 3D convolutional network (E3D), we can achieve state-of-the-art performance over the existing methods on three benchmarks. Besides, we propose a degenerate variant of E3D by replacing 3D convolutions with 2D convolutions and test it on the vehicle datatset TRANCOS, which demonstrates our model can still achieve good results in case the temporal information is not available. 

\bibliography{egbib}
\end{document}